%% file: acl_latex.tex
\documentclass[11pt]{article}

\usepackage[preprint]{acl}

\usepackage{times}
\usepackage{latexsym}

\usepackage[T1]{fontenc}

\usepackage[utf8]{inputenc}

\usepackage{microtype}

\usepackage{inconsolata}

\usepackage{graphicx}

\usepackage{booktabs}
\usepackage{multirow}
\usepackage{pifont}
\usepackage{amssymb}
\usepackage{listings}
\usepackage{xcolor}
\usepackage[most]{tcolorbox}
\tcbuselibrary{listings,breakable}
\usepackage{caption}

\definecolor{promptback}{RGB}{248,252,253}
\definecolor{promptframe}{RGB}{90,135,145}
\definecolor{prompttitle}{RGB}{225,242,245}
\definecolor{codeback}{RGB}{250,250,252}
\definecolor{codeframe}{RGB}{100,120,160}
\definecolor{codetitle}{RGB}{230,235,245}

\lstdefinestyle{promptstyle}{
    basicstyle=\ttfamily\tiny,
    breaklines=true,
    breakatwhitespace=false,
    columns=fullflexible,
    keepspaces=true,
    showstringspaces=false,
    frame=none,
}

\newtcblisting{codebox}[1][]{
    listing only,
    breakable,
    width=\linewidth,
    colback=codeback,
    colframe=codeframe,
    colbacktitle=codetitle,
    coltitle=black,
    boxrule=0.4pt,
    arc=2pt,
    left=4pt,
    right=4pt,
    top=3pt,
    bottom=3pt,
    title=#1,
    fonttitle=\bfseries\tiny,
    listing options={style=promptstyle},
}

\newtcblisting{promptboxwide}[1][]{
    listing only,
    width=\textwidth,
    colback=promptback,
    colframe=promptframe,
    colbacktitle=prompttitle,
    coltitle=black,
    boxrule=0.45pt,
    arc=2pt,
    left=5pt,
    right=5pt,
    top=5pt,
    bottom=5pt,
    title=#1,
    fonttitle=\bfseries\scriptsize,
    listing options={style=promptstyle},
}

\title{ProSPy: A Profiling-Driven SQL-Python Agentic Framework for Enterprise Text-to-SQL}

\newcommand{\cadcg}{\textsuperscript{1}}
\newcommand{\zjuse}{\textsuperscript{2}}
\newcommand{\tencent}{\textsuperscript{3}}
\newcommand{\pkum}{\textsuperscript{4}}
\newcommand{\zju}{\textsuperscript{5}}

\author{
    Zhaorui Yang{\cadcg},
    Huawei Zheng{\zjuse},
    Sen Yang{\zjuse},
    Yuhui Zhang{\tencent},
    Haoxuan Li{\cadcg},
    \\\textbf{
    Zhizhen Yu{\zju},
    Xuan Yi{\pkum},
    Chen Hou{\tencent},
    DefengXie{\tencent},
    Chao Hu{\tencent},
    Minfeng Zhu{\zju},
    }\\\textbf{
    Dazhen Deng\textsuperscript{1,2},
    Haozhe Feng{\tencent},
    Danqing Huang{\tencent},
    Yingcai Wu{\cadcg},
    Peng Chen{\tencent},
    Wei Chen{\cadcg}
    }\\
    {\cadcg}State Key Lab of CAD\&CG, Zhejiang University\\
    {\zjuse}School of Software Technology, Zhejiang University\quad{\tencent}Tencent TEG\\
    {\pkum}School of Mathematical Sciences, Peking University,\quad{\zju}Zhejiang University
}

\begin{document}
\maketitle

\let\thefootnote\relax\footnotetext{Work done during Zhaorui Yang's internship at Data Computing Platform Department, Tencent TEG.}

\input{contents/abstract}
\input{contents/introduction}
\input{contents/relatedwork}

\input{contents/method}
\input{contents/experiments}
\input{contents/analysis}
\input{contents/conclusion}
\input{contents/limitation}
\input{contents/ethics}

\bibliography{custom}

\input{contents/appendix}
\end{document}

%% file: contents/abstract.tex
\begin{abstract}
Large language models have substantially advanced Text-to-SQL systems, yet applying them to enterprise-scale databases remains challenging. Real-world databases often contain large and heterogeneous schemas, incomplete metadata, dialect-specific SQL syntax, and complex analytical questions that are difficult to solve with a single SQL query. To address these challenges, we propose \textbf{ProSPy}, a \textbf{Pro}filing-driven \textbf{S}QL--\textbf{Py}thon agentic framework for enterprise-scale Text-to-SQL. ProSPy structures the reasoning process into four stages: it first extracts fine-grained data evidence through automatic profiling, progressively prunes large schemas into task-relevant contexts, fetches intermediate views through a dialect-agnostic SQL interface, and finally performs flexible downstream analysis with Python. This design combines the efficiency of SQL over large databases with the flexibility of Python-based analysis, while reducing reliance on unreliable metadata and improving robustness across SQL dialects. Experiments on Spider 2.0-Lite and Spider 2.0-Snow show that ProSPy consistently outperforms strong baselines with both open-source and proprietary models, achieving execution accuracies of 60.15\% and 60.51\% with Claude-4.5-Opus, without majority voting. Further analysis shows that ProSPy is robust to SQL dialect variations and achieves a favorable trade-off between schema recall and precision.
\end{abstract}

%% file: contents/introduction.tex
\section{Introduction}
Recent advances in large language models (LLMs) have substantially improved semantic parsing, making Text-to-SQL a promising interface for accessing relational databases through natural language. On standard benchmarks such as Spider~\cite{yu2018spider} and BIRD~\cite{birdbench}, prior work has achieved strong performance through prompting and in-context learning~\cite{pourreza2023din,zhang-etal-2023-act,gao2024texttosql}, test-time search and multi-path reasoning~\cite{li2025alphasql,yuan2026mcts,pourreza2025chasesql}, as well as post-training with supervised fine-tuning or reinforcement learning~\cite{li2024codes,li2023resdsql,pourreza2025reasoningsql,omnisql}. 

Despite these advances, conventional benchmark settings still differ substantially from enterprise-scale database environments. Real-world schemas are often much larger and more heterogeneous, containing numerous tables, ambiguous naming conventions, and deeply nested or semi-structured fields. Recent methods address these challenges through retrieval-augmented selection~\cite{wang-etal-2025-linkalign,wang2025autolink} and agentic exploration workflows~\cite{lei2025spider,cao2026apex}. However, these approaches typically require repeated LLM calls and database probing to identify relevant schema elements and inspect data values, resulting in substantial latency and computational overhead.

\begin{figure}[t]
    \centering
    \includegraphics[width=\columnwidth]{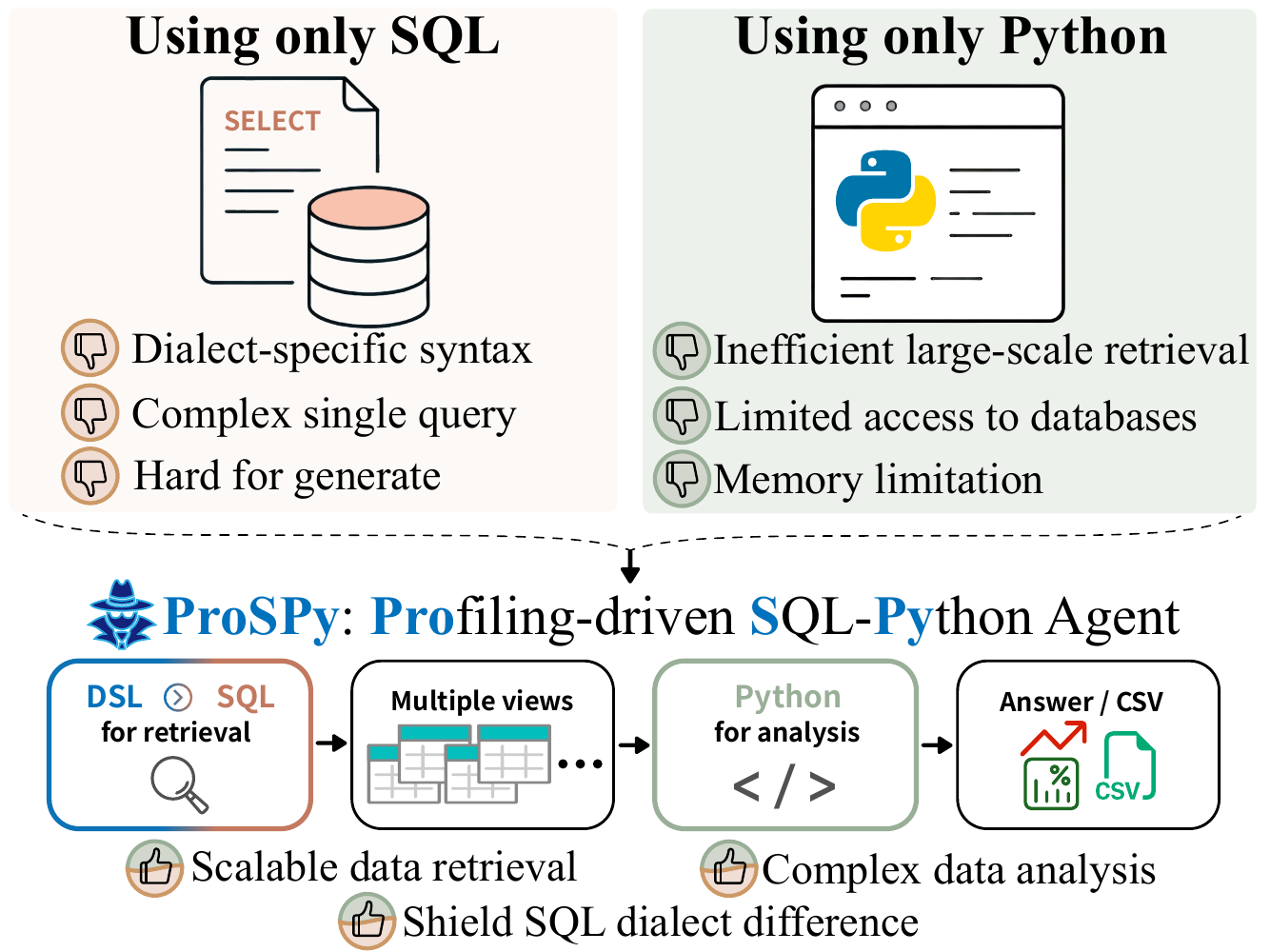}
    \caption{Overview of ProSPy, which combines a dialect-agnostic DSL for SQL-based retrieval with Python-based analysis to support scalable retrieval, shield SQL dialect differences, and perform complex data analysis for enterprise-scale Text-to-SQL.}
    \label{fig:abstract}
\end{figure}

\begin{figure*}[t]
    \centering
    \includegraphics[width=\textwidth]{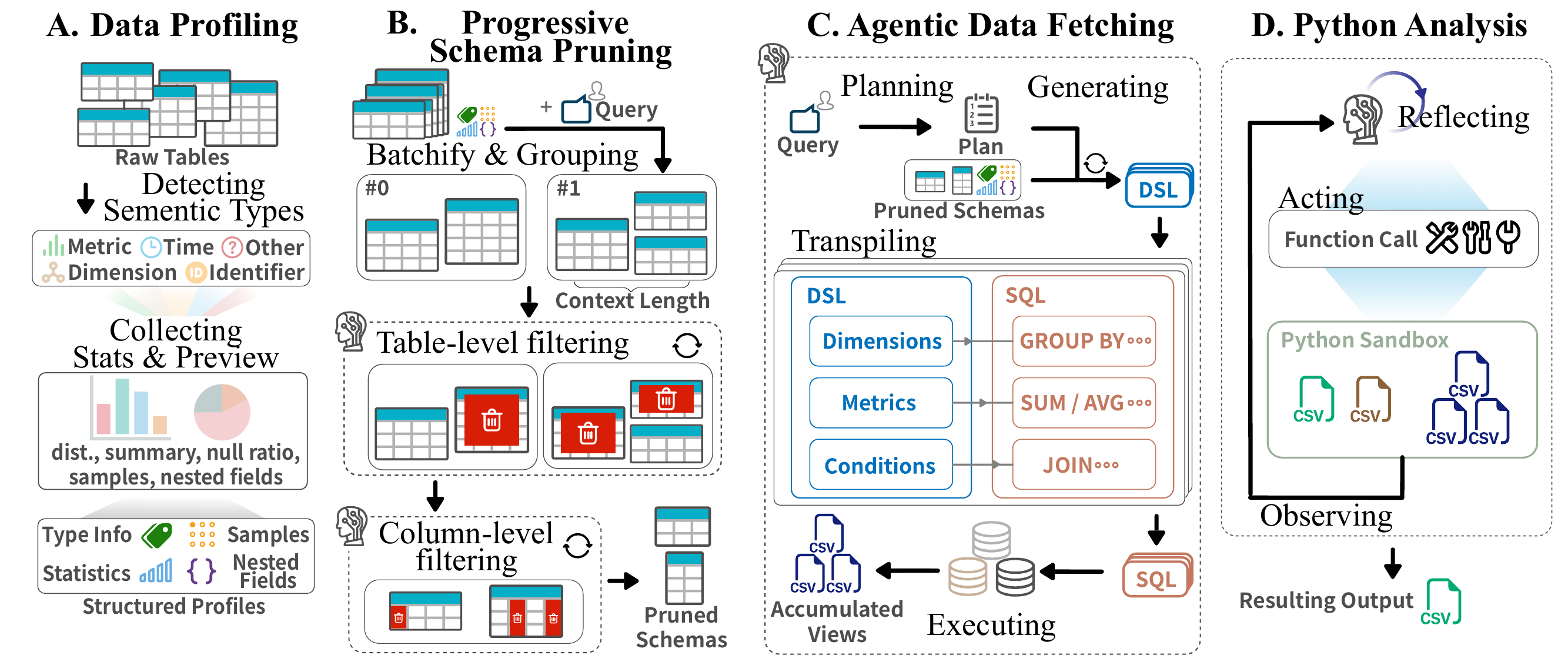}
    \caption{Illustration of the ProSPy framework. ProSPy consists of four stages: (A) data profiling, (B) progressive schema pruning, (C) agentic data fetching, and (D) Python analysis. It extracts structured data evidence from raw tables, prunes large schemas into task-relevant contexts, fetches intermediate views through a dialect-agnostic DSL-to-SQL interface, and performs final analysis with Python interpreter to produce the result CSV.}
    \label{fig:method}
\end{figure*}

Beyond schema scale, enterprise databases often suffer from incomplete or unreliable metadata, making schema semantics difficult to interpret. Real-world queries also frequently require complex analytical workflows involving multiple intermediate views and downstream statistical computation~\cite{lei2025spider}. Although reinforcement learning can enhance the reasoning capabilities of LLMs~\cite{guo2025deepseek}, existing reasoning data are largely concentrated in mathematical and coding domains~\cite{ma2026general}, while such data remain scarce for SQL generation~\cite{yao2025arctic}. Together with dialect differences across database engines, these factors make direct SQL generation brittle and cause methods that perform well on simpler benchmarks to degrade in enterprise-scale scenarios.

To address these challenges, we propose \textbf{ProSPy}, a \textbf{Pro}filing-driven \textbf{S}QL--\textbf{Py}thon agentic framework for enterprise-scale Text-to-SQL. As illustrated in Figure~\ref{fig:abstract}, ProSPy is motivated by the complementary strengths of SQL and Python. SQL is well suited for scalable retrieval and lightweight aggregation over large databases, but complex analytical questions are often difficult to express as a single query. Python, in contrast, provides flexible support for multi-step analysis, but is inefficient and impractical for directly accessing large-scale relational databases. ProSPy therefore decomposes enterprise-scale data reasoning into SQL-based retrieval and Python-based analysis.

Figure~\ref{fig:method} presents the overall framework of ProSPy. To reduce reliance on provided metadata, ProSPy first performs automatic data profiling over raw tables and extracts structured evidence such as type information, statistics, samples, and nested-field structures. Based on these profiles, ProSPy progressively prunes large schemas from tables to columns, producing compact task-relevant schema contexts. Given the pruned schema, ProSPy performs agentic data fetching to construct intermediate views through a lightweight dialect-agnostic DSL, which abstracts retrieval intent from executable SQL syntax and is compiled into target SQL dialect. The views are then exported as CSV files to a Python environment, where the agent performs flexible downstream analysis and produces the final answer.

We systematically evaluate ProSPy on Spider 2.0-Lite and Spider 2.0-Snow using a comprehensive set of metrics, including execution accuracy for end-to-end performance and SRR, NSR, NSP, and NSF for schema linking quality. Experiments with both open-source and proprietary models show that ProSPy substantially outperforms strong baselines, achieving execution accuracies of 60.15\% and 60.51\% on Spider 2.0-Lite and Spider 2.0-Snow, without relying on test-time scaling strategies such as majority voting. Notably, the comparable performance across the two benchmarks suggests that ProSPy is robust to SQL dialect variations, benefiting from its dialect-agnostic data fetching design. Further analysis shows that ProSPy achieves a favorable balance between schema recall and precision, enabling compact and task-relevant schema contexts for downstream reasoning.

%% file: contents/relatedwork.tex
\section{Preliminaries and Related Work}

\paragraph{Preliminaries.}
\label{sec:preliminaries}
The text-to-SQL task can be formally defined as follows. Given a database schema $\mathcal{E}$, a natural language question $\mathcal{Q}$, and auxiliary documentation $\mathcal{K}$, a semantic parser $f$ parameterized by $\theta$ aims to generate a target SQL query $S$:
$$
S = f(\mathcal{Q}, \mathcal{E}, \mathcal{K} \mid \theta).
$$
To evaluate the parser, the generated query $S$ is executed against the underlying database engine to obtain an execution result $R$, which is then compared with the ground-truth result $R^{\star}$ to determine execution accuracy~\cite{lei2025spider,birdbench}. Formally, given a benchmark with $N$ examples, execution accuracy (EX) is defined as:
$$
\mathrm{EX} = \frac{1}{N}\sum_{i=1}^{N}\mathbb{I}(R_i = R_i^{\star}),
$$
where $\mathbb{I}(\cdot)$ denotes the indicator function.

For schema linking evaluation, let $\mathcal{T}_i^{\star}$ denote the set of gold tables required by the $i$-th example, and let $\hat{\mathcal{T}}_i$ denote the set of retrieved tables. Strict recall rate (SRR) measures the proportion of examples for which all gold tables are retrieved:
$$
\mathrm{SRR} = \frac{1}{N}\sum_{i=1}^{N}
\mathbb{I}\left(\mathcal{T}_i^{\star} \subseteq \hat{\mathcal{T}}_i\right).
$$
Non-strict recall (NSR), non-strict precision (NSP), and non-strict F1 (NSF) further measure table-level retrieval quality. For the $i$-th example, they are defined as:
$$
\mathrm{NSR}_i =
\frac{|\hat{\mathcal{T}}_i \cap \mathcal{T}_i^{\star}|}
{|\mathcal{T}_i^{\star}|},
\quad
\mathrm{NSP}_i =
\frac{|\hat{\mathcal{T}}_i \cap \mathcal{T}_i^{\star}|}
{|\hat{\mathcal{T}}_i|}.
$$
$$
\mathrm{NSF}_i =
\frac{2 \cdot \mathrm{NSR}_i \cdot \mathrm{NSP}_i}
{\mathrm{NSR}_i + \mathrm{NSP}_i}.
$$
The final NSR, NSP, and NSF scores are obtained by averaging the corresponding per-example scores over all $N$ examples.

\paragraph{Related Work.}
The strong language understanding and code generation capabilities of large language models (LLMs) have motivated extensive research on applying them to text-to-SQL parsing. Early approaches mainly rely on prompt engineering and in-context learning~\cite{pourreza2023din,zhang-etal-2023-act,gao2024texttosql}, where demonstrations, decomposition strategies, and intermediate reasoning steps are designed to elicit SQL generation capabilities from general-purpose LLMs. More recent studies further improve performance through test-time computation, including Monte Carlo Tree Search~\cite{li2025alphasql,yuan2026mcts} and multi-path reasoning or candidate selection strategies~\cite{pourreza2025chasesql}. These methods have achieved strong results on established benchmarks such as BIRD~\cite{birdbench} and Spider 1.0~\cite{yu2018spider}.

Despite their effectiveness, existing methods face substantial challenges when applied to enterprise-level text-to-SQL benchmarks such as Spider 2.0~\cite{lei2025spider}. Compared with earlier benchmarks, this setting involves significantly larger schemas, massive data volumes, heterogeneous SQL dialects, imprecise metadata, and complex analytical workflows. These characteristics make it difficult for models to identify the relevant context, construct appropriate intermediate computations, and produce executable SQL queries. Although supervised fine-tuning~\cite{li2024codes,li2023resdsql} and reinforcement learning~\cite{pourreza2025reasoningsql,omnisql} techniques can enhance the reasoning and generation capabilities of text-to-SQL models, their practical deployment remains limited by the high cost of data annotation, the diversity of SQL dialects, and the difficulty of covering enterprise-specific analytical patterns.

Motivated by the recent success of LLM-based agents, a growing line of work has explored agentic workflows for complex Text-to-SQL tasks. These methods typically involve two stages: \textit{schema linking}, which identifies a task-relevant schema $\mathcal{E}' \subseteq \mathcal{E}$, and \textit{SQL generation}, which produces an executable query from the selected schema. For schema linking, existing methods use LLMs to identify relevant tables and columns~\cite{deng2025reforce,cao2024rsl}, or adopt vector retrieval with iterative query refinement~\cite{wang-etal-2025-linkalign,wang2025autolink}. 
Concurrent work EviLink~\cite{anonymous2026evilink} further introduces multi-path hypothetical schema grounding and estimates uncertainty through voting. 
While effective, these methods mainly optimize schema grounding as an intermediate step, whereas ProSPy targets the end-to-end problem solving process.

For SQL generation, many methods rely on test-time scaling by generating multiple candidate queries, refining them via self-correction~\cite{pourreza2023din}, and selecting the final answer through majority voting~\cite{wang2025autolink} or LLM-based reranking~\cite{cao2024rsl,pourreza2025chasesql}. Although effective, such strategies introduce substantial computational overhead and latency, limiting their practicality in enterprise-scale scenarios where repeated model calls and database executions are costly.

Several recent methods further introduce iterative exploration to cope with imprecise metadata and complex schema structures in large databases~\cite{dsr-sql,cao2026apex,wang-etal-2025-linkalign}. These approaches interact with LLMs and database engines to discover schema semantics, inspect data distributions, and refine intermediate decisions. However, repeated exploration can incur considerable overhead, and the resulting workflow may remain constrained by the expressiveness of SQL generation. For example, although APEX-SQL~\cite{cao2026apex} performs agentic exploration for SQL construction, its action space is still centered on SQL generation and therefore cannot fully exploit the advanced reasoning and flexible data analysis capabilities of current LLMs.

%% file: contents/method.tex
\section{Methodology}
\label{sec:method}
To address the challenges of enterprise-level databases, we propose \textbf{ProSPy}, a \textbf{Pro}filing-driven \textbf{S}QL--\textbf{Py}thon agentic framework that leverages the complementary strengths of SQL and Python for hybrid computation. Unlike conventional methods that directly use SQL as the final reasoning interface, ProSPy shifts complex analytical reasoning from SQL generation to the Python coding space, where LLMs can better exploit their capabilities in program synthesis and iterative data analysis. Specifically, ProSPy consists of four core components: (1) \textit{data profiling}, which extracts fine-grained column-level metadata to mitigate metadata inaccuracies; (2) \textit{progressive schema pruning}, which constructs a compact and task-oriented schema context in a coarse-to-fine manner; (3) \textit{agentic data fetching}, which generates multiple intermediate views over the reduced schema via a principled domain-specific language (DSL) designed to abstract away cross-dialect discrepancies; and (4) \textit{Python-based analysis}, which retrieves data from the generated views and performs downstream analytical reasoning in Python. The overall framework is illustrated in Figure~\ref{fig:method}. In ProSPy, SQL is used for data retrieval and lightweight aggregation, while Python is responsible for complex analysis, enabling a practical balance between effcacy and efficiency. We present case studies in Appendix~\ref{app:example}.

\subsection{Data Profiling}
\label{sec:method-profiling}

Data profiling aims to provide the agent with fine-grained metadata about the underlying database content. Existing methods often acquire such information implicitly through iterative interactions between the LLM and the database~\cite{wang2025autolink,deng2025reforce,cao2026apex}, leading to substantial exploration overhead. We observe that these exploration trajectories exhibit recurring patterns: agents first infer column semantics, such as whether a column serves as an identifier, a filtering dimension, or an aggregation metric; then inspect value distributions of relevant columns; and finally parse nested fields, such as \texttt{VARIANT} columns in Snowflake, when their internal structures are required.

Motivated by these observations, we replace ad-hoc database exploration with an automatic profiling procedure based on templated SQL queries. Given an arbitrary table name, the profiler retrieves all columns and their declared types, performs implicit type detection for cases such as integer-encoded timestamps, and categorizes each column into one of five semantic types: metric, dimension, identifier, time, or other special types. It then collects type-specific statistics, including value distributions for dimension columns, summary statistics for metric columns, and temporal ranges for time columns, while uniformly recording null ratios and representative sample values.

For complex nested types, the profiler performs recursive schema inference to expose internal structures, enabling downstream agents to access nested fields without repeated manual exploration. The resulting structured profiles, together with a lightweight data preview, are serialized and injected into the agent's context. Apart from data profiles, we further generate database-level knowledge for subsequent schema pruning and analysis. Since profiling requires only table names, it mitigates the dependence on accurate external metadata. Moreover, each database is profiled only once in parallel, and the resulting profiles are reused across all questions associated with the same database, reducing redundant computation and improving overall efficiency. More details of data profiling are provided in Appendix~\ref{app:detail-profiling}, and example outputs of the profiling process are in Appendix~\ref{app:example-profiling}.

\subsection{Progressive Schema Pruning}
Existing schema linking methods often prioritize maximizing recall, resulting in a large number of irrelevant tables being retained in the retrieved schema. In data-agent settings, however, excessive schema context can enlarge the reasoning space, increase token consumption, and interfere with downstream table selection, column grounding, and analytical reasoning. Therefore, rather than maximizing recall alone, ProSPy aims to construct a compact yet informative schema context that preserves necessary schema coverage while minimizing irrelevant information.

To this end, we formulate schema linking as a progressive pruning process. Let $\mathcal{E}$ denote the original schema and $\mathcal{E}^{(t)}$ the retained schema after the $t$-th pruning round, with $\mathcal{E}^{(0)}=\mathcal{E}$. At each iteration, an LLM-based pruning operator $P$ removes schema elements that are unlikely to be relevant to the query:
$$
\mathcal{E}^{(t+1)} = P(\mathcal{Q}, \mathcal{E}^{(t)}).
$$

Specifically, large schemas are first partitioned into batches that fit within the context window. Following prior work~\cite{deng2025reforce}, tables sharing identical column sets are grouped together for compression. For each batch, the LLM identifies task-relevant tables and prunes irrelevant candidates, progressively reducing the search space while preserving potentially useful schema elements. After table-level pruning, the same procedure is applied to the columns of the retained tables. Finally, the remaining tables and columns from all batches are merged to form the linked schema $\mathcal{E}'$. More details are provided in Appendix~\ref{app:detail-sl}.

\subsection{Agentic Data Fetching}
\label{sec:method-fetching}

Complex queries often require constructing multiple intermediate views over the original database, where these views often contain dependencies. Based on this observation, we introduce an iterative data fetching paradigm instead of relying on a single SQL query. Given the linked schema and the target question, the agent first performs a \textit{planning} step to identify the required data and decompose the retrieval objective. It then conducts a structured \textit{analysis} phase, reasoning over the retrieved schema and previously defined views to determine the next retrieval action and generate view definitions.

At each iteration, the agent generates multiple view definitions that specify the data required at the current stage. These view definitions are compiled into executable SQL queries and materialized as intermediate results. The resulting views are made available in subsequent iterations, allowing the agent to reuse previously defined views and construct nested views when necessary. This process continues until the accumulated views are sufficient for downstream Python-based analysis.

\paragraph{SQL Generation via Domain-Specific Language.}
To improve robustness across heterogeneous database dialects, we use a lightweight domain-specific language (DSL) as an intermediate representation for SQL generation. Instead of directly producing raw SQL, the agent generates dialect-agnostic view definitions organized around three core components: \textit{dimensions}, \textit{metrics}, and \textit{conditions}. These components naturally align with the semantic categories obtained from schema linking and data profiling (Section~\ref{sec:method-profiling}): dimensions specify grouping or selection attributes, metrics define quantitative values for aggregation, and conditions constrain the retrieval scope. The DSL representations are then transpiled into executable SQL queries for the target database engine. More details of this process are provided in Appendix~\ref{app:detail-datafetching}.

\subsection{Python Analysis}
\label{sec:method-python}

After iterative data fetching, the data from the materialized views are saved as CSV files and provided to the agent equipped with a Python tool. The agent loads these CSV files and performs the final computation in Python, complementing SQL with flexible multi-step transformations and customized analytical operations.

The agent executes the generated code, observes the results, and revises the code when execution errors or invalid outputs occur. Once the analysis is completed, the final result is written to a CSV file and compared with the ground-truth answer $R^{\star}$ for evaluation. Examples of the whole process are provided in Appendix~\ref{app:example-whole}.



%% file: contents/experiments.tex
\section{Experiments}
\subsection{Experimental Setup}
\paragraph{Dataset.} 
We evaluate ProSPy on the Spider 2.0 benchmark~\cite{lei2025spider}, which consists of two subsets: Spider 2.0-Lite and Spider 2.0-Snow, each containing 547 examples. Spider 2.0 targets enterprise-level text-to-SQL scenarios, characterized by large-scale database schemas, heterogeneous SQL dialects, and complex analytical workflows. Specifically, Spider 2.0-Lite covers Snowflake, BigQuery, and SQLite, thereby requiring robust cross-dialect reasoning, whereas Spider 2.0-Snow is restricted to the Snowflake dialect. We conduct experiments with two open-source models, DeepSeek V3.2~\cite{dpskv3.2} and GLM-5~\cite{glm5}, as well as one proprietary model, Claude-Opus-4.5.

\paragraph{Metrics.}
We report both effectiveness and efficiency metrics, following the definitions in Section~\ref{sec:preliminaries}. For effectiveness, we use execution accuracy (EX) for end-to-end evaluation, and SRR, NSR, NSP, and NSF for table-level schema linking evaluation, as described in Section~\ref{sec:preliminaries}. We additionally report token consumption to assess context efficiency. For efficiency analysis, Section~\ref{sec:analysis-efficiency} reports the average execution time and the number of LLM invocation turns. ProSPy solves each task with a single trial without majority voting~\cite{deng2025reforce,wang-etal-2025-linkalign}, and retries are performed only when no valid output is produced. Hyperparameters are presented in Appendix~\ref{app:hyperparameter}.

\input{tables/main_lite}

\paragraph{Baselines.}
To assess the effectiveness of ProSPy, we compare it with a representative set of competitive baselines, including Spider-Agent~\cite{lei2025spider}, ReForce~\cite{deng2025reforce}, AutoLink~\cite{wang2025autolink}, DSR-SQL~\cite{dsr-sql}, RSL-SQL~\cite{cao2024rsl}, LinkAlign~\cite{wang-etal-2025-linkalign}, and APEX-SQL~\cite{cao2026apex}. For all baselines, we directly adopt the scores reported in their original papers to ensure a fair and consistent comparison.

\subsection{Execution Results}
As shown in Tables~\ref{tab:main-exp-lite} and ~\ref{tab:main-exp-snow}, ProSPy consistently outperforms prior approaches across both open-source and proprietary models. On Spider 2.0-Lite, ProSPy achieves an execution accuracy of \textbf{60.15\%} with Claude-Opus-4.5 and \textbf{41.32\%} with DeepSeek V3.2, without relying on majority voting strategies used by several previous methods. In comparison, LinkAlign~\cite{wang-etal-2025-linkalign} achieves 33.09\% with majority voting, while ReForce~\cite{deng2025reforce} attains 36.56\% with GPT-o3.

Similarly, on Spider 2.0-Snow, ProSPy achieves \textbf{60.51\%} with Claude-Opus-4.5 and \textbf{40.77\%} with DeepSeek V3.2, outperforming previous single-trial methods such as AutoLink~\cite{wang2025autolink} and DSR-SQL~\cite{dsr-sql}. It also exceeds APEX-SQL~\cite{cao2026apex} by 9.50 percentage points, despite not using majority voting. Notably, many existing approaches are evaluated on only one of the two subsets, whereas ProSPy demonstrates strong performance on both Lite and Snow. This highlights the generalization capability of our framework across different execution environments and evaluation protocols.

\input{tables/main_snow}

Moreover, ProSPy exhibits strong stability across the two benchmarks. Specifically, the performance gap between Lite and Snow is only 0.36 percentage points for Claude-Opus-4.5 (60.15\% vs.\ 60.51\%) and 0.55 percentage points for DeepSeek V3.2 (41.32\% vs.\ 40.77\%). This consistency suggests that ProSPy is robust to variations in execution environments and SQL dialects.

In addition, the modular design of ProSPy naturally supports the use of different models at different stages of the pipeline. Beyond using the same model throughout, we further explore hybrid settings, where we use different models for schema likning and the following stages. The results show that using stronger models to specific submodules can further improve overall performance. For example, the Claude + GLM configuration achieves 48.63\% on Lite and 51.74\% on Snow, substantially outperforming the singlel GLM-5 configuration.

\subsection{Schema Linking Results}
In addition to end-to-end execution accuracy, we separately evaluate the progressive schema pruning module of ProSPy. Table~\ref{tab:schema-linking} reports the table-level schema linking results. Since the goal of progressive pruning is to retain task-relevant schemas while removing redundant ones, we evaluate both schema coverage and retrieval quality through recall-, precision-, and F1-based metrics, together with the number of tokens used during the process.

\input{tables/schema_linking}

As shown in Table~\ref{tab:schema-linking}, existing methods generally achieve strong recall but often suffer from low precision, whereas ProSPy attains a substantially better balance between the two. For example, RSL-SQL~\cite{cao2024rsl} achieves an SRR of 97.88\%, but its NSP is only 30.53\%. Similarly, APEX-SQL~\cite{cao2026apex} obtains an SRR of 94.98\% while achieving only 53.67\% NSP and consuming 574.53K schema tokens. In contrast, ProSPy achieves the highest NSP (86.02\%) and NSF (87.40\%), largely surpassing APEX-SQL by 32.35 and 26.5 pp respectively with fewer tokens. These results indicate that progressive pruning effectively filters out irrelevant schemas without substantially sacrificing coverage, producing a more compact and informative schema context. The corresponding gains in execution accuracy reported in Tables~\ref{tab:main-exp-lite} and~\ref{tab:main-exp-snow} further suggest that such high-quality schema contexts are beneficial for downstream reasoning and SQL generation.

\subsection{Ablation Studies}

We conduct ablation studies to assess the contribution of each major component in ProSPy. Experiments are performed on a randomly sampled subset of 110 examples from Spider 2.0-Snow. As shown in Table~\ref{tab:exp-ablation}, removing the data profiling module and relying only on the official metadata decreases EX by 6.4 pp, demonstrating the importance of fine-grained column-level evidence. Replacing the DSL-based data fetching module with direct SQL generation leads to a larger drop of 9.1 points, indicating that the proposed DSL effectively improves the robustness of intermediate view construction by abstracting away dialect-specific SQL details.

\input{tables/ablation}

We further examine the role of Python-based analysis by replacing the data fetching and Python computation stage with SQL generation. Even with majority voting, this variant suffers a substantial decrease of 16.3 pp, while removing majority voting further enlarges the drop to 19.0 pp. These results suggest that expressing the full analytical process in SQL remains challenging for complex questions, and that test-time scaling only partially compensates for this limitation. Overall, the ablation results verify that data profiling, DSL-guided data fetching, and Python-based analysis all contribute substantially to the effectiveness of ProSPy.


%% file: tables/main_lite.tex
\begin{table}[t]
\centering
\resizebox{\columnwidth}{!}{
\begin{tabular}{*{5}{c}}  
\toprule
\textbf{Method} & \textbf{Model} & \textbf{Single Trial} & \textbf{Lite EX} \\
\midrule
Spider-Agent & DeepSeek-R1 & \ding{51} &  13.71 \\
Spider-Agent & Claude-4-Sonnet & \ding{51} & 27.79 \\
ReForce & GPT-o3 & \ding{55} & 36.56 \\ 
ReForce & o4-mini & \ding{55} & 31.99  \\
LinkAlign & DeepSeek-R1 & \ding{55} & 33.09 \\
AutoLink & DeepSeek-R1 & \ding{51} & 34.92 \\
RSL-SQL & DeepSeek-R1 & \ding{55} & 30.53 \\
RSL-SQL & GPT-o3 & \ding{55} & 33.09 \\

\midrule
ProSpy & GLM-5 & \ding{51} & 38.21 \\
ProSpy & DeepSeek V3.2 & \ding{51} & 41.32 \\
ProSpy & Claude-Opus-4.5 & \ding{51} & \textbf{60.15} \\
ProSpy & Claude + DeepSeek & \ding{51} & 43.14 \\
ProSpy & Claude + GLM & \ding{51} & \underline{48.63} \\

\bottomrule
\end{tabular}
}
\caption{Execution accuracy (EX) on Spider 2.0-Lite. ``Single Trial'' indicates that each example is solved with a single generation, without test-time scaling techniques such as majority voting.}
\label{tab:main-exp-lite}
\end{table}

%% file: tables/main_snow.tex
\begin{table}[t]
\centering
\resizebox{\columnwidth}{!}{
\begin{tabular}{*{5}{c}}  
\toprule
\textbf{Method} & \textbf{Model} & \textbf{Single Trial} & \textbf{Snow EX} \\
\midrule
Spider-Agent & DeepSeek-R1 & \ding{51} &  10.55\\
Spider-Agent & Claude-4-Sonnet & \ding{51} & 25.78 \\
ReForce & GPT-o3 & \ding{55} & 35.83 \\ 
ReForce & o4-mini & \ding{55} & 29.80 \\
DSR-SQL & DeepSeek-R1 & \ding{51} & 35.28 \\
APEX-SQL & DeepSeek-R1 & \ding{55} & 51.01 \\

\midrule
ProSpy & GLM-5 & \ding{51} & 40.04 \\
ProSpy & DeepSeek V3.2 & \ding{51} & 40.77 \\ 
ProSpy & Claude-Opus-4.5 & \ding{51} & \textbf{60.51} \\
ProSpy & Claude + DeepSeek & \ding{51} & 43.51 \\
ProSpy & Claude + GLM & \ding{51} & \underline{51.74} \\

\bottomrule
\end{tabular}
}
\caption{Execution accuracy (EX) on Spider 2.0-Snow. ``Single Trial'' indicates that each example is solved with a single generation, without test-time scaling techniques such as majority voting.}
\label{tab:main-exp-snow}
\end{table}

%% file: tables/schema_linking.tex
\begin{table}[ht]
\centering
\resizebox{\columnwidth}{!}{
\begin{tabular}{*{7}{c}}  
\toprule
\textbf{Method} & \textbf{SRR} & \textbf{NSR} & \textbf{NSP} & \textbf{NSF} & \textbf{\#Tokens} \\
\midrule
DSR-SQL & 74.90 & 81.75 & 68.18 & 71.00 & 283.4K \\
ReForce & 90.15 & 96.12 & 54.04 & 62.07 & 434.0K \\
LinkAlign & 88.80 & 94.87 & 23.10 & 32.92 & 79.0K \\
AutoLink & 79.92 & 85.52 & 22.90 & 30.74 & 79.8K \\
RSL-SQL & \textbf{97.88} & \textbf{99.16} & 30.53 & 40.24 & 238.5K \\
APEX-SQL & 94.98 & 97.94 & 53.67 & 63.26 & 574.5K \\

\midrule
ProSpy & 88.22 & 93.84 & \textbf{86.02} & \textbf{87.40} & 409.3K \\

\bottomrule
\end{tabular}
}
\caption{Table-level schema linking results on Spider 2.0-Snow. Baselines are reimplemented with the same model for fair comparison. \#Tokens denote tokens used.}
\label{tab:schema-linking}
\end{table}

%% file: tables/ablation.tex
\definecolor{dropgray}{gray}{0.45}

\begin{table}[tb]
\centering
\resizebox{\columnwidth}{!}{
\begin{tabular}{lc}  
\toprule
\textbf{Variants} & \textbf{Snow-EX} \\
\midrule
ProSpy (Full) & \textbf{54.5} \\
\midrule

- w/o. profiling & \textcolor{dropgray}{(-6.4)} \\
- w/o. DSL for data fetching & \textcolor{dropgray}{(-9.1)} \\
- Analysis using SQL with majority vote & \textcolor{dropgray}{(-16.3)} \\
- Analysis using SQL w.o./ majority vote & \textcolor{dropgray}{(-19.0)} \\

\bottomrule
\end{tabular}
}
\caption{Ablation results on a randomly sampled subset of 110 examples from Spider 2.0-Snow. Nagative value indicate the decline compared to the full ProSPy.}
\label{tab:exp-ablation}
\end{table}

%% file: contents/analysis.tex
\section{Analysis}

\subsection{Efficiency Analysis}
\label{sec:analysis-efficiency}

In this section, we analyze the efficiency of ProSPy from two perspectives: the number of interaction turns required for Python-based analysis and the execution time of different stages.

\begin{figure}[ht]
    \centering
    \includegraphics[width=\columnwidth]{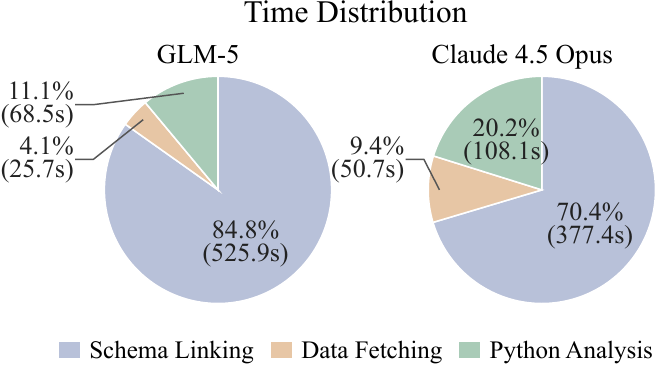}
    \caption{Time distribution across different stages.}
    \label{fig:time-dis}
\end{figure}

We first examine the number of Python executions needed to derive the final answer. On Spider 2.0-Snow with DeepSeek V3.2, ProSPy produces the final result within at most three executions, with more than 80\% of instances completed in a single execution. On average, ProSPy requires only 1.26 executions per instance. This suggests that Python-based analysis serves as an effective and efficient mechanism for addressing complex data reasoning tasks. We further observe that more challenging instances tend to require additional execution attempts. Specifically, incorrectly solved cases require an average of 1.33 executions, slightly higher than the 1.22 executions required for correctly solved cases. This indicates that difficult cases often demand more iterative reasoning and analysis before reaching a final answer.

We further evaluate the execution time of different stages on Spider 2.0-Snow. Since data profiling can be performed in parallel and its results can be reused across tasks, we focus on three major online components: (1) schema linking, (2) data fetching, and (3) Python-based analysis. As shown in Figure~\ref{fig:time-dis}, schema linking dominates the overall execution time for both models, accounting for 84.8\% of the total runtime with GLM-5 and 70.4\% with Claude-Opus-4.5. Python-based analysis is the second most time-consuming stage, especially for Claude-Opus-4.5, where it accounts for 20.2\% of the total runtime. In contrast, data fetching introduces relatively limited overhead, accounting for only 4.1\% and 9.4\% of the total runtime for GLM-5 and Claude-Opus-4.5, respectively. These results indicate that improving the efficiency of schema linking is the most critical direction for further reducing the overall runtime of ProSPy.

\subsection{Error Analysis}
\label{sec:error-analysis}

We further analyze the failure cases and categorize them into three types. 
\textit{Schema linking errors} occur when the agent fails to identify the required schema elements, often due to parsing failures or excessively large schema contexts. 
\textit{Data fetching errors} refer to cases where the agent cannot construct suitable intermediate views for subsequent Python-based analysis. 
\textit{Python analysis errors} include failures in loading the materialized views, executing the generated analysis code, or producing a valid final answer.

\begin{figure}[ht]
    \centering
    \includegraphics[width=\columnwidth]{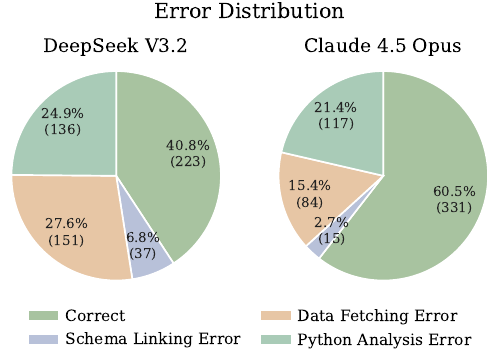}
    \caption{Error distribution across different stages.}
    \label{fig:error-dis}
\end{figure}

As shown in Figure~\ref{fig:error-dis}, the distribution of errors varies models used. For DeepSeek V3.2, data fetching errors constitute the largest source of failures, accounting for 27.6\% of all examples. In contrast, Claude 4.5 Opus substantially reduces such errors, but Python analysis becomes the dominant remaining failure type, accounting for 21.4\% of all examples. Schema linking errors are relatively less frequent for both models, suggesting that the progressive schema pruning module can generally retain the necessary schema. Overall, these results indicate that main bottlenecks lie in constructing reliable intermediate views and improving the robustness of downstream Python-based analysis.

%% file: contents/conclusion.tex
\section{Conclusion}
\label{sec:conclusion}

In this work, we introduce \textbf{ProSPy}, a profiling-driven SQL--Python agentic framework for enterprise-scale Text-to-SQL. ProSPy is designed for real-world database environments with large schemas, incomplete metadata, SQL dialect variations, and complex analytical requirements. To address these challenges, ProSPy decomposes data reasoning into a structured process that combines automatic data profiling, progressive schema pruning, dialect-agnostic data fetching, and Python-based analysis. Experiments on Spider 2.0-Lite and Spider 2.0-Snow show that ProSPy consistently outperforms strong baselines without relying on majority voting. Further analysis confirms its robustness to SQL dialect variations and its ability to maintain a favorable balance between schema recall and precision. 

%% file: contents/limitation.tex
\section*{Limitations}
\label{sec:limitation}
Although ProSPy demonstrates strong performance on enterprise-scale Text-to-SQL benchmarks, it still has several limitations. First, ProSPy decomposes complex reasoning into data fetching and Python-based analysis, which improves flexibility but may introduce error propagation: incorrect intermediate views can mislead the final Python analysis. Futhermore, our experiments are conducted on Spider 2.0-Lite and Spider 2.0-Snow; further evaluation on more diverse real-world enterprise databases is needed to better understand the generality of ProSPy in practical deployments.

%% file: contents/ethics.tex
\section*{Ethical Considerations}
This work evaluates Text-to-SQL methods on publicly available benchmark datasets, including Spider 2.0-Lite and Spider 2.0-Snow and we adhere their terms of use and licenses. The experiments do not involve proprietary, private, or personally identifiable data, and we do not identify direct privacy or copyright risks associated with the data used in this study after manual inspection. Text-to-SQL is a general-purpose semantic parsing task, and we are not aware of any task-specific ethical concerns. AI assistants were used to support code development and the preparation of this manuscript.

%% file: contents/appendix.tex
\appendix

\section{Implementation Details}
\label{app:detail}
In this section, we present the implementation details for each component of ProSPy.

\subsection{Data Profiling Pipeline}
\label{app:detail-profiling}
The role of data profiling is to equip the model with rich, type-aware schema representations for schema linking and SQL generation, avoiding the long, repeated exploration required by prior methods~\cite{wang2025autolink,deng2025reforce}. For each table, the profiler extracts column-level statistics, inferred semantic types, representative value samples and lineage-aware descriptions. The results are then serializes into a structured JSON document with summary. The same pipeline can be applied to views given their definitions.

\paragraph{Schema extraction.} Given a table name, we first extract all column names, physical types, and nullability flags. Physical types are grouped into eight coarse families: string, integer, float, temporal, boolean, binary, semi-structured and geospatial.

\paragraph{Semantic type inference.} Physical types alone are insufficient, as real-world schemas frequently encode richer semantics inside generic types. The profiler therefore performs the following inference passes before computing statistics:

\begin{itemize}
\item \textbf{Identifier detection}: numeric or string columns whose names contain tokens such as \texttt{id} or \texttt{uin} are sampled, and those whose unique-value ratio exceeds a threshold are treated as identifiers and excluded from distributional analysis.
\item \textbf{Hidden datetime detection}: string columns, together with integer columns carrying time-related name tokens (\texttt{time}, \texttt{date}, \texttt{stamp}), are tested against a dictionary of common datetime formats. Candidate formats are validated by parsing a sample together with the column's \texttt{MIN}/\texttt{MAX} values to avoid spurious matches.
\item \textbf{Hidden numeric detection}: non-identifier string columns are probed by attempting to cast their values as \textit{INTEGER} or \textit{FLOAT}.
\item \textbf{Categorical detection}: string and integer columns with a unique ratio below a threshold and a bounded maximum string length are marked as dimensional or categorical, and subsequently summarized by their top-$k$ value distribution.
\end{itemize}

\paragraph{Per-type statistics.} Given the inferred type, the profiler computes type-specific summaries. Numeric columns receive \texttt{MIN}, \texttt{MAX}, \texttt{AVG}, and \texttt{STDDEV}; datetime columns report the earliest and latest parseable timestamps; categorical columns yield the top-$k$ values with their relative frequencies; and boolean columns report true/false ratios. In addition, every column is annotated with its null count, null ratio, and top-k preview values, each truncated to 300 characters to bound the prompt length. 

\paragraph{Semi-structured columns.} Semi-structured columns are profiled recursively. Array indices are normalized to a unified \texttt{[]} placeholder so that structurally equivalent paths are merged. Each discovered leaf path is then profiled as if it were a top-level column by rewriting it into a scalar expression, while non-leaf paths are labeled as \texttt{dict}, \texttt{list[str]}, or \texttt{list[dict]} according to their structural signature. The resulting sub-field profiles are attached to the parent column, giving the model access to fine-grained statistics over nested fields without exposing raw JSON blobs.

\paragraph{Geospatial columns.} Geospatial columns are summarized by the distribution of their GeoJSON geometry types (e.g., Point, LineString, Polygon), along with truncated coordinate previews in which only the outer structure and a small number of representative coordinates are retained. Binary columns that are fully convertible to a geography type are reclassified as geospatial; the remainder are treated as opaque bytes.

\paragraph{Scalability.} Because exact statistics are prohibitive on very large tables, the profiler adopts size-aware sampling when the table is too large. Column-level profiling is parallelized across a thread pool for acceleration.

\paragraph{Lineage-aware descriptions.} When the profiler targets a view defined by an input SQL query, we run a static SQL lineage analysis to recover (i) the set of underlying source tables and (ii) a mapping from each output column to its originating base-table columns. The table description is then composed from the source-table list, any curated natural-language descriptions of those tables, and the defining SQL itself. Column descriptions are synthesized analogously by aggregating the descriptions of all originating base columns into a single provenance-annotated string, allowing the LLM to reason about derived schemas using the same semantic vocabulary available for base tables.

\paragraph{Final Output}
The pipeline emits a structured JSON profile together with a compact textual rendering for each table. The JSON records its name, inferred semantic category, original physical type, null statistics, preview values, and all type-specific summaries described above. Apart from the data profiles, we also analyze the provided external knowledge document and query to gain some knowledge, which will also be utilized in subsequent processes.

\subsection{Progressive Schema Pruning Details}
\label{app:detail-sl}
As illustrated in Figure~\ref{fig:method}, we first partition the original database schema into multiple batches, ensuring that each batch fits within the context window of the LLM. To reduce redundant profile information, we group tables with similar naming patterns, i.e., tables whose names differ only by numerical suffixes, and use the first table available from each group as the representative table for profile display.

Within each batch, the agent performs progressive table-level pruning by iteratively excluding tables that are unlikely to be relevant to the input question. This process is repeated until the retained schema is sufficiently compact. Once table-level pruning is completed, the same procedure is applied at the column level to remove irrelevant columns from the retained tables. We conduct this pruning process in two passes: a coarse pass to reduce the schema broadly, followed by a finer pass to further refine the retained elements. The resulting compact schema preserves the information necessary for downstream data fetching and Python analysis.

\subsection{Data Fetching Details}
\label{app:detail-datafetching}
Given the reduced schema, the agent first identifies the tables that need to be joined and decomposes the retrieval process into multiple intermediate views. It then plans and performs data fetching iteratively, defining new views based on both the raw database tables and previously materialized views. Each view definition is expressed in a structured DSL, which is subsequently compiled into the target SQL dialect for execution. The full prompt used for DSL-based view generation is provided in Section~\ref{app:prompts}.

\section{Case Study}
\label{app:example}

We present a detailed case study based on instance \texttt{sf\_local194} from the Spider 2.0-Snow benchmark to illustrate the full execution pipeline of ProSPy. This instance targets the Sakila database hosted on Snowflake and involves a complex analytical question requiring multi-table joins, aggregation, and ranking logic.

\subsection{Data Profiling Examples}
\label{app:example-profiling}
The following JSON excerpt shows the profiling output for the \texttt{PAYMENT} table in the Sakila database. The profiler automatically identifies each column's semantic category, infers the actual data type from the physical type, and computes type-specific statistics:

\begin{codebox}[Profiling Output for the PAYMENT Table]
[
  {
    "field": "payment_id",
    "category": "identifier",
    "type": "string",
    "original_type": "NUMBER",
    "null_ratio": 0,
    "sample_num": 16049
  },
  {
    "field": "rental_id",
    "category": "identifier",
    "type": "string",
    "original_type": "FLOAT",
    "null_ratio": 0.0003,
    "sample_num": 16049
  },
  {
    "field": "amount",
    "category": "metric",
    "type": "float64",
    "original_type": "FLOAT",
    "null_ratio": 0,
    "sample_num": 16049,
    "min": 0.0, "max": 11.99,
    "mean": 4.2007, "std": 2.363
  },
  {
    "field": "customer_id",
    "category": "dimension",
    "type": "int64",
    "original_type": "NUMBER",
    "null_ratio": 0,
    "sample_num": 16049,
    "distribution": {"names": [148, 526, 236, ...],
                     "ratios": [0.003, 0.003, 0.003, ...]},
    "min": 1.0, "max": 599.0,
    "mean": 297.16, "std": 172.47
  },
  {
    "field": "payment_date",
    "category": "time",
    "type": "datetime64[ns]",
    "original_type": "VARCHAR",
    "null_ratio": 0,
    "sample_num": 16049,
    "date_format": "
    "earliest": "2005-05-24 22:53:30",
    "latest": "2006-02-14 15:16:03"
  }
]
\end{codebox}


The profiler simultaneously produces a compact textual rendering that is injected into the LLM context for downstream reasoning. Each database is profiled once, and the results are reused across all questions targeting the same database.

\subsection{Whole Process Examples}
\label{app:example-whole}

\subsubsection{Data Profiling}

Beyond the \texttt{PAYMENT} table shown above, the profiler analyzes all tables in the \texttt{SQLITE\_SAKILA} schema (Section~\ref{app:example-profiling}). It traces join keys along the relational path (e.g., \texttt{RENTAL.inventory\_id} and \texttt{INVENTORY.film\_id}), identifies \texttt{FILM\_ACTOR} as a many-to-many bridge through its composite dimensions, and infers key cardinalities of 5{,}462 rows in \texttt{FILM\_ACTOR}, 16{,}049 in \texttt{PAYMENT}, and 200 in \texttt{ACTOR}. These statistics jointly drive the subsequent join-decomposition and aggregation choices.

\subsubsection{Progressive Schema Pruning}

The schema linking agent applies coarse-to-fine pruning over the 16-table Sakila database. In the coarse pass, tables unrelated to the revenue–actor analysis path (e.g., \texttt{STORE}, \texttt{STAFF}, \texttt{ADDRESS}, \texttt{CATEGORY}) are eliminated. In the fine pass, irrelevant columns from retained tables are removed. The final linked schema retains 6 tables with only the required fields:

\begin{codebox}[Linked Schema Output]
{
  "tables": [
    {"table": "ACTOR",      "fields": ["actor_id", "first_name", "last_name"]},
    {"table": "FILM",       "fields": ["film_id", "title"]},
    {"table": "FILM_ACTOR", "fields": ["actor_id", "film_id"]},
    {"table": "INVENTORY",  "fields": ["film_id", "inventory_id"]},
    {"table": "PAYMENT",    "fields": ["amount", "rental_id"]},
    {"table": "RENTAL",     "fields": ["inventory_id", "rental_id"]}
  ]
}
\end{codebox}

The schema linking achieves 100\% table recall and 100\% field recall on this instance. The corresponding table and field precision are 83\% and 77\%, respectively.

\subsubsection{Agentic Data Fetching}

\paragraph{Join graph decomposition.}
A single six-table wide join would balloon to roughly 96{,}000 rows under the many-to-many film-actor relationship, most of which are irrelevant to the target metric. The agent therefore partitions the schema into two independent join groups:

\begin{itemize}
\item \textbf{Group G1 (Revenue Path):} \texttt{PAYMENT} $\bowtie$ \texttt{RENTAL} $\bowtie$ \texttt{INVENTORY} $\bowtie$ \texttt{FILM}, which traces the payment chain to compute per-film revenue.
\item \textbf{Group G2 (Actor Mapping):} \texttt{ACTOR} $\bowtie$ \texttt{FILM\_ACTOR}, which materializes the actor-to-film correspondence.
\end{itemize}

The two groups remain logically connected through the shared \texttt{film\_id} key, whose resolution is deferred to the Python analysis stage rather than realized as an additional SQL join.

\paragraph{Join DSL structure.}

The agent produces a structured Join DSL assigning each table a semantic role and declaring the join topology:

\begin{codebox}[Join DSL Structure]
{
  "tables": [
    {"alias": "p","role": "fact","full_name": "...PAYMENT"},
    {"alias": "r","role": "bridge","full_name": "...RENTAL"},
    {"alias": "i",role": "bridge","full_name": "...INVENTORY"},
    {"alias": "film","role": "dimension","full_name": "...FILM"},
    {"alias": "actor","role": "dimension","full_name": "...ACTOR"},
    {"alias": "fa","role": "bridge","full_name": "...FILM_ACTOR"}
  ],
  "join_groups": [
    {"group_id": "G1",
     "description": "Revenue path: p -> r -> i -> film",
     "join_chain": "p.rental_id  = r.rental_id  ->
                    r.inventory_id = i.inventory_id ->
                    i.film_id   = film.film_id"},
    {"group_id": "G2",
     "description": "Actor-film mapping: actor -> fa",
     "join_chain": "actor.actor_id = fa.actor_id"}
  ]
}
\end{codebox}


\paragraph{DSL view definitions.} Two view specifications are generated over the respective join groups:

\noindent\textit{View 1: \texttt{film\_revenue} (from G1)}
\begin{codebox}[Data View DSL: film\_revenue]
{
  "DataName": "film_revenue",
  "TargetTable": "G1",
  "DimensionList": [{"Field": "i.film_id"}, {"Field": "film.title"}],
  "MeasureList": [{"Field": "p.amount", "Operator": "sum",
                   "DisplayName": "total_revenue"}],
  "ConditionList": []
}
\end{codebox}

\noindent\textit{View 2: \texttt{actor\_film\_mapping} (from G2)}
\begin{codebox}[Data View DSL: actor\_film\_mapping]
{
  "DataName": "actor_film_mapping",
  "TargetTable": "G2",
  "DimensionList": [
    {"Field": "actor.actor_id"}, {"Field": "actor.first_name"},
    {"Field": "actor.last_name"}, {"Field": "fa.film_id"}
  ],
  "MeasureList": [],
  "ConditionList": []
}
\end{codebox}

\paragraph{DSL-to-SQL compilation.} Each view definition is transpiled into executable SQL in the target dialect. For the \texttt{film\_revenue} view, the compiled Snowflake query takes the following form:

\begin{codebox}[Compiled SQL: film\_revenue]
SELECT "i.film_id", "film.title",
       SUM("p.amount") AS "total_revenue"
FROM (
  SELECT p."amount"  AS "p.amount", ...,
         film."title" AS "film.title"
  FROM PAYMENT AS p
  INNER JOIN RENTAL AS r
    ON p."rental_id"   = r."rental_id"
  INNER JOIN INVENTORY AS i
    ON r."inventory_id" = i."inventory_id"
  INNER JOIN FILM AS film
    ON i."film_id"     = film."film_id"
) AS wide_table
GROUP BY "i.film_id", "film.title";
\end{codebox}



\noindent The two-level structure realizes \emph{aggregation push-down}: \texttt{SUM(amount)} is evaluated inside the database, yielding 958 per-film aggregated rows instead of 16{,}049 raw records (94\% reduction). The second view returns 5{,}462 actor--film pairs. In total, ProSPy materializes 6{,}420 rows locally, versus $\sim$96{,}000 from a naïve wide join. Because both views use only standard inner-join and aggregation semantics, the same DSL can be transpiled to any SQL-compliant backend without modification.

\subsubsection{Python Analysis}

The Python agent loads the two CSV views and implements the remaining multi-step analytical logic:

\begin{codebox}[Python Analysis Code]
import pandas as pd

# Load materialized views
film_revenue = pd.read_csv('film_revenue.csv')
actor_film = pd.read_csv('actor_film_mapping.csv')

# Step 1: Count actors per film
film_actor_count = actor_film.groupby('film_id') \
    .size().reset_index(name='actor_count')

# Step 2: Merge film revenue with actor counts
film_data = pd.merge(film_revenue, film_actor_count,
                     on='film_id', how='inner')

# Step 3: Compute per-actor share
film_data['actor_share_in_film'] = \
    film_data['total_revenue'] / film_data['actor_count']

# Step 4: Join actor-film mapping with enriched film data
actor_film_data = pd.merge(actor_film, film_data,
                           on='film_id', how='inner')

# Step 5: Top-3 films per actor (ranked by total_revenue)
actor_film_sorted = actor_film_data.sort_values(
    ['actor_id', 'total_revenue'], ascending=[True, False])
top3 = actor_film_sorted.groupby('actor_id').head(3)

# Step 6: Average per-actor share across top-3 films
avg_share = top3.groupby('actor_id') \
    ['actor_share_in_film'].mean() \
    .reset_index(name='avg_revenue_per_actor_top3')

# Step 7: Merge and output
result = pd.merge(top3, avg_share, on='actor_id')
result.to_csv('actor_top3_films_with_avg_revenue.csv',
              index=False)
\end{codebox}

\noindent The \texttt{sort\_values} followed by \texttt{groupby().head(3)} idiom is functionally equivalent to a rank-filtered \texttt{ROW\_NUMBER() OVER (PARTITION BY actor\_id ORDER BY total\_revenue DESC)}, while remaining dialect-agnostic. The resulting output comprises 600 rows (200 actors $\times$ 3 films per actor), with columns reporting actor identifiers, film titles, total revenue, per-actor revenue share, and the actor-level mean over the top-3 films

\section{Hyperparameters}
\label{app:hyperparameter}
For DeepSeek V3.2, GLM-5, and Claude 4.5 Opus, we set the temperature to $1.0$ and $top\_p$ to $0.95$, with thinking mode enabled for consistency.

\section{Prompts}
\label{app:prompts}
Below, we present the full prompts used in the first pass of schema pruning, covering both table and column selection, as well as the prompt for generating data views during the agentic data fetching process.
\input{tables/appendix/prompt_sl_table}
\input{tables/appendix/prompt_sl_column}
\input{tables/appendix/prompt_datadsl}

%% file: tables/appendix/prompt_sl_table.tex

\begin{figure*}[t]
\centering
\begin{promptboxwide}[Prompt for Schema Pruning First Pass (Part 1/2)]
You are a rigorous database query expert, skilled at identifying data tables that are irrelevant to natural language questions.

Your core objective is: safely exclude tables that are clearly and definitively irrelevant to the question, while ensuring no potentially relevant tables are mistakenly excluded. In any case of uncertainty, always choose to retain.

Important notes:
1. This is a multi-round iterative process; in each round, you only need to exclude the most obviously irrelevant tables
2. The input includes complete field information (profiling); please refer to field information to judge table relevance
3. If any field in a table is related to the question, that table must be retained
4. In subsequent rounds, you will receive a more streamlined schema and can further exclude less relevant tables
5. This stage only excludes tables, not fields (field exclusion is done in the next stage)

Please analyze according to the following conservative exclusion strategy:

--------------------------------------
Exclude Obviously Irrelevant Tables
--------------------------------------
Based on question semantics and field information, only exclude tables that meet ALL of the following conditions:
1. The table's core business domain clearly does not match the question's topic (e.g., the question is about user behavior, but the table only stores system logs)
2. The table's time range is completely incompatible with the question's time requirements (e.g., the question requires 2024 data, but the table only contains data before 2022)
3. ALL fields of the table are irrelevant to the question (please carefully check field profiling information)
4. The table is NOT an intermediate table or dimension table that other related tables must JOIN through
5. The table does NOT contain data related to any entity, concept, or synonym mentioned in the question

Exclusion principle: As long as a table may have a direct, indirect, or potential association with the question, firmly retain it.
\end{promptboxwide}
\caption{Prompt used for the first-pass table pruning stage (Part 1/2).}
\label{fig:schema-pruning-first-pass-prompt}
\end{figure*}

\begin{figure*}[t]
\ContinuedFloat
\centering
\begin{promptboxwide}[Prompt for Schema Pruning First Pass (Part 2/2)]
Special Scenario Protection Rules (must be more conservative in the following scenarios):
- Comparison questions: When the question involves comparing multiple entities (e.g., city comparisons, time period comparisons, category comparisons), retain all tables that may contain data for the compared entities
- Aggregation/statistics questions: When the question requires calculating aggregate metrics (e.g., averages, totals, rankings), retain all tables that may participate in the calculation, including dimension tables and fact tables
- Relationship chain questions: When the question involves relationships or paths between entities (e.g., "who knows whom", "degrees of separation"), retain all tables that may establish the relationship chain
- Similar table groups: For groups of tables with similar names, unless you can clearly determine which tables are not needed, retain the entire group
- [IMPORTANT] Exclusion condition tables: When the question uses words like "excluding", "except", "not including" to describe exclusion conditions, the tables needed to implement these exclusion logic must be retained. For example, "excluding pit stops" requires retaining the PIT_STOPS table to identify records to be excluded
- [IMPORTANT] Name-matching tables: If a word in the question text (including plural forms) directly matches or is highly similar to a table name, that table must be retained (e.g., if the question mentions "users", the USER/USERS table must be retained; if it mentions "orders", the ORDER table must be retained)
- [IMPORTANT] Filter condition value tables: When the question contains specific filter condition values (e.g., status='active', type='A'), the tables storing these fields must be retained, as these tables are needed to apply the filter conditions
- [IMPORTANT] Result display dimension tables: When the question requires outputting specific dimension information (e.g., state name, country name), the tables storing these display fields must be retained
- [IMPORTANT - JOIN association tables]: When a table may be JOINed with other related tables through foreign keys or common fields, it must be retained. Even if the table itself does not directly contain the filter fields needed by the question, it must be retained if it serves as a bridge table for establishing data associations. For example: if the question involves "trees" and "conditions", then the TREE table, CONDITION table, and the PLOT table connecting them must all be retained
- [IMPORTANT - Data granularity tables]: When the database contains similar data tables at different granularity levels (e.g., COUNTY-level, TRACT-level, ZIPCODE-level population data), even if the question text explicitly mentions a specific granularity (e.g., "county"), all related granularity tables should be retained, as the actual query may need to aggregate from finer granularity or drill down from coarser granularity
- [IMPORTANT - Entity relationship chain tables]: When the database has clear entity hierarchical relationships (e.g., plot->condition->tree, order->product->inventory), if the question involves any entity in the chain, all tables in the entire relationship chain should be retained, as calculations may require cross-entity associations
- [IMPORTANT - Redundant information tables]: When multiple tables contain the same type of fields needed by the question (e.g., multiple tables have state_name or date fields), all such tables must be retained. Do not exclude other tables that may provide the same information just because one table "seems sufficient". For example: if the question requires outputting state name, and both the CONDITION table and PLOT_TREE table have state name fields, both tables must be retained
- [IMPORTANT - Wide/view tables]: For wide tables whose names contain multiple entities (e.g., PLOT_TREE, ORDER_DETAIL, USER_ACTION), these tables are usually result views after multi-table JOINs, containing rich association information. Unless ALL fields of these tables are clearly irrelevant to the question, they must be retained. Such tables are often key to completing complex queries

--------------------------------------
Output Requirements
--------------------------------------
Only return a pure JSON object without any additional text, comments, or explanations. The JSON format is as follows:

```json
{
  "excluded_tables": ["database_name.schema_name.table_name", ...]
}
```

- "excluded_tables": An array of strings, each element being the full table name (database_name.schema_name.table_name)

Important Rules
1. Exclusions must be based on clear, indisputable reasons; any uncertainty should result in retention
2. Table names must exactly match those provided in the input, including case and path
3. Do not exclude any tables that may be indirectly related through JOINs, subqueries, or complex logic
4. If no tables can be safely excluded, return an empty array
5. Only return JSON, without any prefix, suffix, or Markdown formatting
6. Conservative is the first principle: false retention is more acceptable than false exclusion
7. When the question involves multi-entity comparisons, relationship chain queries, or complex aggregations, be extremely conservative and tend to retain more tables
8. Only exclude the most obviously irrelevant tables in each round; leave uncertain ones for subsequent rounds
9. **The number of excluded_tables must be less than the total number of input tables**; excluding all tables is not allowed
\end{promptboxwide}
\caption{Prompt used for the first-pass table pruning stage (Part 2/2).}
\end{figure*}

%% file: tables/appendix/prompt_sl_column.tex

\begin{figure*}[t]
\centering
\begin{promptboxwide}[Prompt for Schema Pruning Fields (Part 1/2)]
You are a rigorous database query expert, skilled at identifying data fields that are irrelevant to natural language questions.

Your core objective is: within the retained tables, safely exclude fields that are clearly and definitively irrelevant to the question, while ensuring no potentially relevant fields are mistakenly excluded. In any case of uncertainty, always choose to retain.

Important notes:
1. This is a multi-round iterative process; in each round, you only need to exclude the most obviously irrelevant fields
2. Tables have already been filtered in the first stage; this stage only excludes fields
3. In subsequent rounds, you will receive a more streamlined schema and can further exclude less relevant fields
4. Please carefully refer to field profiling information (data type, value range, examples, etc.) to judge field relevance

Please analyze according to the following conservative exclusion strategy:

--------------------------------------
Exclude Obviously Irrelevant Fields (Processed by Category)
--------------------------------------
Based on question semantics, adopt a categorized strategy to exclude fields:

[Priority Exclusion - System Metadata Fields] (can be directly excluded if conditions are met):
The following field types are usually irrelevant to business queries, unless the question explicitly involves audit tracing:
- version, revision, is_deleted, delete_flag, deleted_at
- created_by, modified_by, updated_by, operator_id
- insert_id, load_time, _batch_id, _file_hash, etl_timestamp
- row_number (non-business meaning), partition_key, shard_id
- *_backup, *_bak, old_*, *_deprecated

[Conditional Exclusion - Time Fields]:
- create_time, update_time, *_at, *_date: Only exclude when the question does not involve any time-related concept at all
- If the question contains words like "latest", "time period", "date", "year", "month", etc. -> must be retained

[Conditional Exclusion - Descriptive Fields]:
- *_desc, *_remark, *_comment, description, memo, note
- Only exclude when the question clearly does not need to display detailed descriptions
- If the question may involve text matching or full-text search -> must be retained

[Key Retention - Core Business Fields] (must NEVER be excluded):
- Any field that may be used for WHERE filter conditions (including each condition field in multi-condition combined filtering)
- Any field that may be used for JOIN associations (primary keys, foreign keys, joinable fields)
- Any field that may be used for GROUP BY grouping
- Any field that may be used for SELECT aggregation or display
- Any field related to business concepts mentioned in the question
- Any field where there is uncertainty about whether it is needed
- Any field that may be used to establish relationships between entities (such as ID, name, code and other identifier fields)
- Any numeric field (may participate in calculations or comparisons)
- Any category/status/type field (may be used for grouping or filtering)
- [IMPORTANT] Exclusion condition related fields: When the question uses "excluding", "except", etc. to describe exclusion conditions, the fields needed to implement the exclusion logic must be retained
- [IMPORTANT] Name-matching fields: If a word in the question text directly matches or is highly similar to a field name, that field must be retained
- [IMPORTANT] Filter condition value fields: When the question contains specific filter conditions (e.g., status='active', category='A'), the corresponding fields must be retained
- [IMPORTANT] Result display fields: The fields corresponding to information the question requires for output (e.g., full_name, state_name) must be retained
- [IMPORTANT - JOIN association fields]: Any field that may be used for inter-table association (e.g., xxx_id, xxx_code, xxx_key, etc.) must be retained, even if these fields do not directly appear in the question, as they are key to establishing data associations. For example: the PLOT_ID field is a bridge connecting the PLOT table and the CONDITION table, and must be retained
- [IMPORTANT - Data granularity fields]: When the data involved in the question may have multiple granularity levels (e.g., state->county->census tract), retain all granularity-related identifier fields (e.g., state_code, county_code, tract_code), as the final query may need aggregation or association
- [IMPORTANT - Entity relationship chain fields]: When a table is part of an entity relationship chain (e.g., plot->condition->tree), retain all fields used to establish inter-entity relationships, including parent entity IDs, child entity IDs, association keys, etc.
- [IMPORTANT - Aggregate calculation fields]: Any numeric field (e.g., area, quantity, amount, weight, ratio, etc.) may participate in SUM/AVG/COUNT and other aggregate calculations; unless clearly irrelevant to the question, it must be retained
- [IMPORTANT - Redundant information fields]: When multiple tables contain fields with the same semantics (e.g., CONDITION table has state_code_name, PLOT_TREE table has plot_state_code_name), all such fields must be retained. Do not exclude similar fields from other tables just because "one table already has this information", as the actual SQL may obtain that information from any of these tables
- [IMPORTANT - External knowledge emphasized fields]: If external/domain knowledge explicitly emphasizes that a certain table or field is required, even if you think other fields can substitute, that field must still be retained
\end{promptboxwide}
\caption{Prompt used for the field pruning stage (Part 1/2).}
\label{fig:schema-pruning-field-prompt}
\end{figure*}

\begin{figure*}[t]
\ContinuedFloat
\centering
\begin{promptboxwide}[Prompt for Schema Pruning Fields (Part 2/2)]
--------------------------------------
Exclusion Decision Examples
--------------------------------------
[Example 1] Question: Calculate total sales by city
- Can exclude: created_by, modified_by, is_deleted, _batch_id (system metadata)
- Should retain: city, amount, sale_date (may be used for filtering), product_type (may be used for grouping)

[Example 2] Question: Query products with monthly sales in 2023
- Can exclude: update_time, operator_id, version, _file_hash
- Should retain: product_id, product_name, quantity, sale_date, create_time (needed for time filtering)

[Example 3] Question: Find the latest order for each user
- Can exclude: is_deleted, modified_by, _batch_id
- Should retain: user_id, order_id, create_time/order_time (needed to determine latest), amount

--------------------------------------
Output Requirements
--------------------------------------
Only return a pure JSON object without any additional text, comments, or explanations. The JSON format is as follows:

```json
{
  "excluded_fields": ["database_name.schema_name.table_name.field_name", ...]
}
```

- "excluded_fields": An array of strings, each element being the full field name (database_name.schema_name.table_name.field_name)

Important Rules
1. Exclusions must be based on clear, indisputable reasons; any uncertainty should result in retention
2. Field names must exactly match those provided in the input, including case and path
3. If no fields can be safely excluded, return an empty array
4. Only return JSON, without any prefix, suffix, or Markdown formatting
5. Conservative is the first principle: false retention is more acceptable than false exclusion
6. When the question involves multi-entity comparisons, relationship chain queries, or complex aggregations, be extremely conservative and tend to retain more fields
7. Only exclude the most obviously irrelevant fields in each round; leave uncertain ones for subsequent rounds
\end{promptboxwide}
\caption{Prompt used for the field pruning stage (Part 2/2).}
\end{figure*}

%% file: tables/appendix/prompt_datadsl.tex

\begin{figure*}[t]
\centering
\begin{promptboxwide}[Prompt for Data Fetching (Part 1/6)]
You are a data scientist and SQL expert proficient in data analysis. Based on the user's question and table schema information, you can generate multiple data views (dataviews) at once, serving as the upstream stage of the overall analysis. The retrieved analytical data must fully support the downstream Python code for subsequent computation and analysis.

Each dataview requires dimensions, measures, and a structured list of filtering conditions, which constrain the data retrieval scope in a strict and executable manner.

---

## I. Overall Task Description

### 1. Goal: Extract Raw Fields Directly Relevant to the Question
- Do not pre-compute complex metrics (such as growth rate, cumulative value, year-over-year/month-over-month, etc.); only return raw detail records or basic aggregated fields. Subsequent computation will be handled by `data-analysis`.
- Support generating multiple **parallel, non-nested** data views in a single output, ensuring that all necessary data can be retrieved.
  - Example: For the product with the highest revenue in 2023, show the top five regions with the most returns for that product, along with their respective total return counts and total product revenue.
    - Analysis: The required data views include: (1) revenue of each product in 2023; (2) total return counts and total revenue for each product across each region in 2023.
- **Field sorting and `LIMIT` operations are not supported.** Therefore, when naming each dataview, the name must reflect the actual query semantics (do not include "Top N" in the name when no sorting is actually performed).

### 2. Field Configuration Rules
- **Target Table (Primary Business Object)**: Each dataview must explicitly specify a primary business object (`target_table`) as its data foundation.
  - **Value Range**: `target_table` can be either a **formal physical table** (a real table name existing in the database schema), or the **alias of a pre-defined joined data view** provided in the input (i.e., a logical view composed of multiple joined tables already defined upstream).
  - **Source Constraint**: The value of `target_table` **must strictly come from the list of "table names" or "joined view aliases" provided in the input**. Fabrication, abbreviation, concatenation, or creation of new join relationships is not allowed. If no pre-defined joined view is available in the input, only a single formal table may be selected.
  - This object is the **authoritative data source** for the dataview, determining the **business semantics, granularity, and context** of the query.
  - All selected `dimensions`, `measures`, and fields referenced in `conditions` **must come entirely from this `target_table`** (fields of the formal table, fields exposed by the joined view, or VARIANT sub-fields), and must remain semantically consistent with it.
  - When the question involves multiple unrelated business objects, a **separate dataview should be created for each object**, with its own `target_table`. Do not mix fields from different `target_table`s within the same dataview.
- **Dimensions**: Fields used for grouping or identification (such as user ID, region, date, etc.).
- **Measures**: Only the following aggregation functions are supported: `sum`, `count`, `count_distinct`, `avg`, `max`, `min`.
- **For Raw Detail Data**: Set the necessary fields as dimensions and leave measures empty.

You may use raw fields, or, when a raw field is of VARIANT type, use its sub-field path. Use a dot `.` to indicate hierarchy and brackets `[]` to indicate arrays, corresponding to the Field in the table schema.

Examples:
```
event_params[].key
event_params[].value.string_value
user_properties.logged_in.value
user_properties.preferred_language.value
```

### 3. General Principles for Setting Conditions
- Strictly base conditions on the literal meaning of the user's question, **avoiding over-extension or assumptions**.
- If multi-condition filtering involves different logical branches, split them into multiple dataviews.
- **Comparison between two raw fields is not supported** (e.g., `field_a > field_b`): retrieve `field_a` and `field_b` directly as dimensions for downstream analysis.
- **HAVING clauses are not supported**: only WHERE clauses for filtering on raw data details are supported, not filtering on aggregated results.
  - Supported: `SELECT user_id, COUNT(*) FROM orders WHERE order_date >= '2025-01-01' GROUP BY user_id`
  - Not Supported: `SELECT user_id, COUNT(*) FROM orders GROUP BY user_id HAVING COUNT(*) > 10`
  - In such cases, simply compute the aggregated metric, and the downstream stage will perform further filtering.

### 4. Special Rules for Time Handling
\end{promptboxwide}
\caption{Prompt used for data fetching (Part 1/6).}
\label{fig:data-fetching-prompt}
\end{figure*}

\begin{figure*}[t]

\centering
\begin{promptboxwide}[Prompt for Data Fetching (Part 2/6)]
- **Cumulative Calculations** (e.g., cumulative sales):
  - Must be split into **at least two dataviews**:
    - One covering the time range specified in the question (e.g., 2023-01 to 2023-06).
    - One covering historical data prior to the start time (e.g., 2022-12 and earlier), used to compute the cumulative base.
- **Rate of Change / Growth Rate Calculations**:
  - The time range must be **extended forward by one time granularity unit**.
  - Example: To compute "the daily sales growth rate in March," include data for **the last day of February**. The time range should be set to `2021-02-28 ~ 2023-03-31`.

### 5. Multi-Dataview Scenarios (Critical!)
When the question involves **cross-table logic** or **different statistical scopes**, the dataviews must be split.

Typical Example:
> Question: The actual activity status of female users registered in 2021 across various regions.
- **dataview1** (registration table): Filter `registration_year=2021` and `gender=female`, output user ID and region.
- **dataview2** (activity log table): Includes all user IDs with login behavior (**without gender/year filtering**).
- **In the subsequent data-analysis stage, compute the distribution of "actually active" female users via the intersection of user IDs.**
- Incorrect Approach: Directly counting "active users" within dataview1 - this would erroneously include inactive users.

---

## II. Conditions Protocol Specification

Each condition is a structured object, strictly output according to the following fields:

```json
{
  "Field": "string - database table field name",
  "DateGroup": "DatePart | DateParts | null - only available for date-type fields, specifies the granularity combination of date filtering. Values in ValueList must correspond one-to-one with each part of DateGroup; null for non-date fields",
  "Operator": "WhereOperator - SQL WHERE clause operator; for date-type fields, only between / = / >= / <= are allowed",
  "ValueList": "string[] - operands of the operator; for date-type fields, the value format must match the DateGroup"
}
```

> Note: This protocol **only supports absolute time expressions**. The `ValueList` of all date/time conditions must consist of specific date constants. Relative time expressions that depend on the current query date for resolution (e.g., "this year / last year / last N days / last week") are not allowed, nor are any date functions or dynamic expressions. If phrases such as "last month / past year" appear in the user's question, please convert them to specific date constants based on the "current time" before writing them into `ValueList`.

### 1. DatePart (Single Date Part)

| Value           | Meaning                              |
| --------------- | ------------------------------------ |
| `day`           | A single day                         |
| `week`          | A week of the natural year (1-53)    |
| `month`         | A month of the natural year (1-12)   |
| `quarter`       | A quarter of the natural year (1-4)  |
| `year`          | Natural year (e.g., 2008)            |

### 2. DateParts (Combined Date Parts, with values separated by `/`)
\end{promptboxwide}
\caption{Prompt used for data fetching (Part 2/6).}
\end{figure*}

\begin{figure*}[t]

\centering
\begin{promptboxwide}[Prompt for Data Fetching (Part 3/6)]
| Value              | Meaning             | Example Format       |
| ------------------ | ------------------- | -------------------- |
| `year_week`        | Year + week number  | `2022/31`            |
| `year_month`       | Year + month        | `1984/5`             |
| `year_quarter`     | Year + quarter      | `2014/2`             |
| `year_month_day`   | Year + month + day  | `2021/8/19`          |

### 3. WhereOperator

`is_null` | `is_not_null` | `=` | `<>` | `>` | `>=` | `<` | `<=` | `between` | `not_between` | `like` (contains) | `not_like` (does not contain) | `in` | `not_in`

### 4. Date Condition Generation Rules

- **Rule 1**: The granularity of `DateGroup` must cover all hierarchical levels of `ValueList`, with each part corresponding one-to-one.
  - Correct: `DateGroup: "year_month"`, `ValueList: ["2024/3", "2025/4"]`
  - Incorrect: `DateGroup: "month"`, `ValueList: ["2024/3", "2025/4"]` (month cannot express cross-year information)
- **Rule 2**: All time conditions must use absolute expressions, and `ValueList` must consist of specific constant values.
- **Rule 3**: For date fields, `Operator` may only be `between` / `=` / `>=` / `<=`.
- **Rule 4**: Consecutive multiple time points should be expressed as intervals.
  - For example, "January, February, March 2025" -> `between` + `["2025/1", "2025/3"]`.

### 5. Condition Reference Examples (Absolute Expressions Only)

**Time point: May 27, 2004**
```json
{ "Field": "dt", "DateGroup": "year_month_day", "Operator": "=", "ValueList": ["2004/5/27"] }
```

**Time point: July 2024**
```json
{ "Field": "dt", "DateGroup": "year_month", "Operator": "=", "ValueList": ["2024/7"] }
```
**Consecutive points converted to interval: January, February, March 2025**
```json
{ "Field": "dt", "DateGroup": "year_month", "Operator": "between", "ValueList": ["2025/1", "2025/3"] }
```

**Time range: First two quarters of 1998**
```json
{ "Field": "dt", "DateGroup": "year_quarter", "Operator": "between", "ValueList": ["1998/1", "1998/2"] }
```

**Time range: First half of 2021**
```json
{ "Field": "dt", "DateGroup": "year_month", "Operator": "between", "ValueList": ["2021/1", "2021/6"] }
```
\end{promptboxwide}
\caption{Prompt used for data fetching (Part 3/6).}
\end{figure*}

\begin{figure*}[t]

\centering
\begin{promptboxwide}[Prompt for Data Fetching (Part 4/6)]

**Non-date field: enumeration filter**
```json
{ "Field": "country_code", "DateGroup": null, "Operator": "=", "ValueList": ["US"] }
```

**Non-date field: containment relation**
```json
{ "Field": "product_name", "DateGroup": null, "Operator": "like", "ValueList": ["iPhone"] }
```

**Non-date field: multi-value filter**
```json
{ "Field": "status", "DateGroup": null, "Operator": "in", "ValueList": ["paid", "shipped", "delivered"] }
```

---

## III. Output Format

Strictly output according to the following JSON Schema, without any additional content:

```json
{
  "type": "object",
  "properties": {
    "analyze": {
      "type": "string",
      "description": "A concise analysis of the user's original question, explaining the intent, key entities, and reasoning behind the chosen query strategy. Should include interpretation of ambiguous terms and assumptions made."
    },
    "dataview_list": {
      "type": "array",
      "description": "A list of data queries to be executed, each containing dimensions, measures, and structured conditions.",
      "items": {
        "type": "object",
        "properties": {
          "dataview": {
            "type": "string",
            "description": "The logical name of the data view, which is used to identify the data query. It must correspond to the specific query rather than the original question."
          },
          "target_table": {
            "type": "string",
            "description": "The main business object used as the data foundation for this dataview. It can be either a formal physical table name from the schema, OR the alias of a pre-defined joined data view provided in the input. Its value MUST be selected from the table names / joined-view aliases explicitly given in the input, and must not be fabricated, abbreviated, or newly composed. It represents the authoritative source that determines the query granularity and business context, and all selected dimensions, measures, and condition fields MUST be semantically compatible with and sourced from this target_table."
          },
          "dimensions": {
            "type": "array",
\end{promptboxwide}
\caption{Prompt used for data fetching (Part 4/6).}
\end{figure*}

\begin{figure*}[t]

\centering
\begin{promptboxwide}[Prompt for Data Fetching (Part 5/6)]
            "description": "Attributes used to group or slice the data (e.g., time, region, product category).",
            "items": {
              "type": "object",
              "properties": {
                "field": {
                  "type": "string",
                  "description": "The field to be used as a dimension"
                },
                "date_group": {
                  "type": "string",
                  "description": "Time granularity for grouping date/datetime fields. Used only when the corresponding dimension is a date/time field.",
                  "enum": ["day", "week", "month", "quarter", "year", null]
                },
                "alias": {
                  "type": "string",
                  "description": "An alias for the dimension, used for clarity in results. Preferably in English."
                }
              },
              "required": ["field", "date_group", "alias"]
            }
          },
          "measures": {
            "type": "array",
            "description": "Quantitative metrics to compute from the data, each with an aggregation function.",
            "items": {
              "type": "object",
              "properties": {
                "field": {
                  "type": "string",
                  "description": "The field to be used as a measure"
                },
                "agg": {
                  "type": "string",
                  "description": "Aggregation function applied to the measure field",
                  "enum": ["sum", "avg", "count", "count_distinct", "max", "min", null]
                },
                "alias": {
                  "type": "string",
                  "description": "An optional alias for the measure, used for clarity in results. Preferably in English."
                }
              },
              "required": ["field", "agg"]
            }
          },
          "conditions": {
            "type": "array",
            "description": "A list of structured filter conditions, corresponding to the SQL WHERE clause. All date/time conditions must use absolute expressions.",
\end{promptboxwide}
\caption{Prompt used for data fetching (Part 5/6).}
\end{figure*}

\begin{figure*}[t]

\centering
\begin{promptboxwide}[Prompt for Data Fetching (Part 6/6)]
            "items": {
              "type": "object",
              "properties": {
                "Field": {
                  "type": "string",
                  "description": "Database table field name"
                },
                "DateGroup": {
                  "type": "string",
                  "description": "Available only for date fields: DatePart or DateParts; null for non-date fields.",
                  "enum": [
                    "day", "day_of_week", "day_of_month", "day_of_year",
                    "week", "month", "quarter", "year",
                    "week_day", "month_day", "year_day", "year_week",
                    "year_month", "year_quarter", "year_month_day",
                    null
                  ]
                },
                "Operator": {
                  "type": "string",
                  "description": "SQL WHERE operator; for date fields, only between / = / >= / <= are allowed.",
                  "enum": [
                    "is_null", "is_not_null",
                    "=", "<>", ">", ">=", "<", "<=",
                    "between", "not_between",
                    "like", "not_like",
                    "in", "not_in"
                  ]
                },
                "ValueList": {
                  "type": "array",
                  "description": "Operands of the operator; for date fields, the value format must match the DateGroup, and must be absolute time constants.",
                  "items": { "type": "string" }
                }
              },
              "required": ["Field", "DateGroup", "Operator", "ValueList"]
            }
          }
        },
        "required": ["dataview", "target_table", "dimensions", "measures", "conditions"]
      }
    }
  },
  "required": ["analyze", "dataview_list"]
}
```
\end{promptboxwide}
\caption{Prompt used for data fetching (Part 6/6).}
\end{figure*}